\documentclass{article}

\newcommand{\LoGAssetRoot}{}
\IfFileExists{LoG/log_2026.sty}{%
  \renewcommand{\LoGAssetRoot}{LoG/}%
}{}
\makeatletter
\edef\input@path{{\LoGAssetRoot}{./}}
\makeatother

\PassOptionsToPackage{hypertexnames=false}{hyperref}
\usepackage{log_2026}
\makeatletter
\renewcommand{\@noticestring}{Accepted at LoG 2026.}
\makeatother
\usepackage[sort,round]{natbib}

\usepackage{url}
\usepackage{booktabs}
\usepackage{multirow}
\usepackage{amsfonts}
\usepackage{amsmath}
\usepackage{amssymb}
\usepackage{graphicx}
\usepackage{wrapfig}
\usepackage{caption}
\usepackage{needspace}
\usepackage{placeins}
\usepackage{xcolor}
\usepackage{array}
\usepackage{algorithm}
\usepackage{algpseudocode}

\IfFileExists{figures/iclr2027_fisher_final.pdf}{%
  \graphicspath{{figures/}}%
}{%
  \graphicspath{{../figures/}}%
}

\newcommand{\rell}{\ensuremath{\mathrm{rel}L_2}}

\newcommand{\px}[1]{\partial_{x}#1}
\newcommand{\pxx}[1]{\partial_{xx}#1}
\newcommand{\pt}[1]{\partial_{t}#1}
\providecommand{\CDR}{\textsc{cdr}}

\title[Amortizing Physics-Informed Neural Solvers via Graph Hypernetworks]{Amortizing Physics-Informed Neural Solvers\\
via Graph Hypernetworks}

\author[C. Jing et al.]{%
Cheng Jing\thanks{Work done during internship at Applied Materials Inc.}\\
Arizona State University\\
\texttt{cjing6@asu.edu}\And
Abhishek Verma\\
Applied Materials Inc.\\
\texttt{AbhishekKumar\_Verma@amat.com}\And
Kallol Bera\\
Applied Materials Inc.\\
\texttt{Kallol\_Bera@amat.com}\And
Yixuan He\\
Arizona State University\\
\texttt{Yixuan.He@asu.edu}\And
Kookjin Lee\thanks{Corresponding author.}\\
Arizona State University\\
\texttt{Kookjin.Lee@asu.edu}
}

\begin{document}

\maketitle

\begin{abstract}
Amortizing physics-informed neural networks (PINNs) across related PDEs requires describing each equation
to a reusable solver. Coefficient vectors encode numerical parameters in predefined slots, leaving operator
and cross-field assignments implicit. We make these relationships explicit in an operator graph, with
nodes for fields, derivatives, terms, and residuals and coefficients retained as term attributes.
A graph hypernetwork generates diagonal codes that initialize a meta-trained factorized PINN for each
target equation. Meta-training and target-specific adaptation use governing equations and prescribed
conditions without solution labels. We compare coefficient-vector, DeepSets-based term-set, and graph
conditioning by solution accuracy within a fixed adaptation budget. In scalar
convection--diffusion--reaction problems, both term-based descriptors improve high-reaction accuracy,
with similar performance. In two-field Fisher--KPP, meta-training sees uncoupled and one-way systems;
after 3,000 adaptation steps on unseen two-way coupling, the graph's mean final error is $35.7\%$ below
the term set and $67.7\%$ below the coefficient vector. In a fixed-structure capacitively coupled plasma
model, the coefficient vector performs best. These results support extending coefficient conditioning
with explicit equation relationships for physics-based solver adaptation.
\end{abstract}

\section{Introduction}
\label{sec:intro}
Scientific modeling often requires solving a family of related partial differential equations (PDEs) rather than
a single equation. Physics-informed neural networks (PINNs) construct solutions by optimizing the governing
residuals and prescribed conditions \citep{raissi2019physics}, but repeating this optimization for every instance
can be costly. Hypernetwork-based amortization shares information across instances by predicting
equation-specific solver configurations that can be further adapted using physics residuals
\citep{de2021hyperpinn,cho2023}. This raises a representation question: \emph{what information about an equation
should the hypernetwork receive?}

Coefficient-conditioned hypernetworks describe each instance through numerical parameters in a predefined
equation layout. Such layouts can cover different term combinations and coupling directions by reserving
equation-specific slots for candidate terms. Our focus is how these relationships are presented to the
hypernetwork: implicitly through slot conventions or explicitly through connections. Coupled PDEs
make this distinction concrete. The strength of an interaction and the variables it connects are both part of
the equation's meaning: a source $\kappa v$ in the equation for $u$ and a source $\kappa u$ in the equation for
$v$ describe opposite coupling directions, even when $\kappa$ is unchanged. This motivates supplying the
hypernetwork with differential operators and cross-field relationships alongside their numerical coefficients.

We represent this information as an \emph{operator graph}: fields, derivatives, additive terms, and equation
residuals are nodes, and their connections specify how the PDE is assembled. Coefficients remain attributes
of the terms, while connectivity makes operator composition and coupling explicit. A graph hypernetwork
reads this representation and predicts per-layer configuration codes for a shared factorized PINN backbone
\citep{cho2023}. The predicted codes and meta-trained backbone form an equation-conditioned initialization.
The target solver is then adapted by minimizing its physics residual and enforcing its prescribed conditions.
This combines coefficients and connectivity within a shared operator vocabulary; both training phases use
physics objectives without solution labels.

We evaluate this approach on three PDE families, each with one spatial dimension and its own trained model.
Cross-structure convection--diffusion--reaction (\CDR{}) tests combinations of familiar terms, including a
high-reaction regime with a documented PINN optimization failure mode \citep{krishnapriyan2021}.
Our two-field Fisher--KPP variant builds on linearly coupled reaction--diffusion systems \citep{holzer2016proof}
and tests generalization from uncoupled and one-way-coupled systems to previously unseen two-way coupling.
A simplified capacitively coupled plasma (CCP) model \citep{jing2026extending} tests application to a
continuity--Poisson system whose coupling structure stays fixed while its coefficients vary. Comparisons with coefficient-vector and
DeepSets-based term-set baselines \citep{zaheer2017deep} show benefits from term-based conditioning in the
high-reaction \CDR{} regime and lower graph-model error on unseen Fisher--KPP coupling. The coefficient vector
performs best in fixed-structure CCP. This contrast is consistent with the hypothesis that the value of
relational conditioning depends on what varies within the equation family, rather than on coupling alone.

Our contributions are:
\begin{itemize}\itemsep2pt
\item \textbf{Operator-graph conditioning for amortized PINNs (\S\ref{sec:method}).} A graph hypernetwork
  configures a shared neural solver from a representation that combines numerical coefficients, differential
  operators, and cross-field relationships.
\item \textbf{Evidence across term and coupling variation (\S\ref{sec:exp}).} Under fixed target-adaptation
  budgets, term-based conditioning improves high-reaction \CDR{} accuracy, and graph conditioning gives
  the lowest mean error on unseen Fisher--KPP coupling composition. A fixed-structure continuity--Poisson
  comparison identifies a setting where coefficient conditioning performs best.
\end{itemize}

\section{Background}
\label{sec:bg}
We review the PINN solver, its factorized backbone, and the hypernetwork and message passing used to configure it.

\paragraph{Physics-informed neural networks.} Let $u(x,t)$ denote a PDE solution and $u_\theta(x,t)$ its
neural approximation, with solver parameters $\theta$. For a coupled system, $u$ collects the fields $u_f$ and
$\mathcal{R}$ collects their equation residuals; parameter subscripts are suppressed in architectural formulas.
Given $\mathcal{R}[u]=0$, physics-informed training minimizes
$\tfrac{1}{|X|}\sum_{(x,t)\in X}\|\mathcal{R}[u_\theta](x,t)\|_2^2$ over space--time collocation points $X$,
together with initial and boundary conditions \citep{raissi2019physics}. These conditions are imposed either
through loss penalties or directly in the network output. Derivatives are computed by automatic differentiation.
Repeated optimization across PDE instances motivates amortization, especially when the physics loss is
difficult to optimize \citep{krishnapriyan2021}.

\paragraph{Factorized PINN backbone.} We adopt the layer parameterization of \citet{cho2023} for each field. For field $f$
and hidden layer $\ell$, the configured weight is
$W_f^\ell=C^\ell\,\mathrm{diag}(s_f^\ell)\,R^\ell$, where
$C^\ell,R^\ell\in\mathbb{R}^{w\times w}$ are shared learned bases and $w$ is the hidden width.
The code $s_f^\ell\in\mathbb{R}^{w}$ specifies the effective diagonal for that instance and field.
With these full-width bases, the hypernetwork configures each layer through $w$ diagonal entries rather than
predicting all $w^2$ matrix entries. The shared bases are learned across the PDE family; the codes are
instance-specific.

\paragraph{Hypernetworks.} A hypernetwork $H$ predicts parameters of another neural network \citep{ha2017}.
In Hyper-LR-PINN, PDE coefficients condition the diagonal codes of a factorized solver \citep{cho2023}.
We retain this code interface, $s=H(d)$ with $s=\{s_f^\ell\}$, and compare a coefficient vector, a set of
terms, and an operator graph as the descriptor $d$. The predicted codes configure
the shared PINN, providing an initial approximation that is subsequently adapted using the physics objective.
This shares the cost of learning useful solver configurations across instances; the training and deployment
phases are described in \S\ref{sec:method}.

\paragraph{Message passing.} On a graph $G=(V,E)$, node $i\in V$ has a hidden state $h_i^k$ at round $k$.
A message-passing update combines that state with its neighbors' states using shared learned functions
\citep{gilmer2017neural}. After $K$ rounds, information can propagate across a $K$-hop neighborhood.
A readout pools node states into a graph-level or field-specific representation. Section~\ref{sec:method}
specifies how this representation configures the PINN.

\section{Method}
\label{sec:method}
We learn an equation-conditioned initialization for physics-based adaptation. A hypernetwork reads the
PDE's operator graph and predicts diagonal codes; these codes and the meta-trained backbone initialize
the target PINN. We then optimize its codes and backbone using the target physics objective. The prediction
before adaptation is the \emph{initial approximation}. We describe the graph, configured solver, and
optimization phases below.

\subsection{Operator graph}
For conditioning, we express each governing residual as additive terms, each represented by a scalar
coefficient and its field or derivative factors. These terms define the operator graph $G=(V,E)$.
The physics loss evaluates the same governing equations by automatic differentiation; graph construction
and loss evaluation may use algebraically equivalent residual forms (App.~\ref{app:graph}).
The graph has four node kinds: fields, derivatives, terms, and equation residuals. Connections link a field to
its derivatives, each factor to the terms containing it, and each term to its target residual
(App.~\ref{app:graph}).

Node features $z_i$ record the node kind and its attributes: field identity and derivative orders where
applicable, and coefficient sign, log-magnitude, and arity (number of factors) for each nonzero term.
For example, $-\kappa v$ in the equation for $u$ gives the path $v\to(-\kappa v)\to\mathcal{R}_u$:
the connections identify the source field and target equation, while the term node retains the coupling
coefficient. Field identities and distinct field/residual node kinds retain these source and target roles
even when message passing uses symmetrized connections. Changing terms or their assignments changes the
graph over this shared vocabulary. Coefficients and connectivity therefore describe complementary parts
of the same equation.

\subsection{Graph hypernetwork and PINN}
The hypernetwork $H$ embeds node features as $h_i^0=\phi(z_i)$, aggregates neighboring states, and decodes
per-field configuration codes. The Fisher--KPP implementation uses
\[
  h_i^{k+1}=h_i^k+\sigma\!\Big(W_{\text{self}}h_i^k+
  W_{\text{neigh}}\!\textstyle\sum_{j\in V}\hat a_{ij}h_j^k\Big),
\]
where $\sigma$ is GELU and $W_{\text{self}},W_{\text{neigh}}$ are shared across nodes and rounds.
The weights $\hat a_{ij}$ come from symmetrizing graph connections, adding self-loops, and normalizing
each row. All edges use one neighbor matrix; coefficient signs enter through node features.
After $K$ rounds ($K=2$ in the main comparison), a readout for field $f$ with node $i_f$ predicts codes for the $L$ factorized layers:
\begin{equation}
  r_f=\rho\big(\big[\,h_{i_f}^{K}\,;\,\bar h\,\big]\big),\qquad
  s_f^{\ell}=\mathrm{ReLU}\!\big(A^{\ell}r_f+b^{\ell}\big)\in\mathbb{R}^{w}\quad(\ell=1,\dots,L),
  \label{eq:codes}
\end{equation}
where $\phi$ is the learned node embedding, $\rho$ is the learned readout network, $[\,;\,]$ denotes
concatenation, and $\bar h$ is the mean of the final term and field states. The learned $A^\ell,b^\ell$
decode layer $\ell$. Family-specific aggregation and readout settings are given in
App.~\ref{app:descriptors}. Appendix~\ref{app:fisher-depth} varies the propagation depth with
training and test depths matched.

The codes configure field $f$'s MLP $g_f$ through
$W_f^\ell=C^\ell\,\mathrm{diag}(s_f^\ell)\,R^\ell$ (\S\ref{sec:bg}): $H$ predicts diagonal codes rather
than full weight matrices. With a fixed coordinate encoding $\gamma$ and output transform $\mathcal{A}_f$,
the prediction is $u_f(x,t)=\mathcal{A}_f\!\big(g_f(\gamma(x,t)),x,t\big)$.
Fisher--KPP uses periodic spatial features and a hard initial-condition transform; CCP uses RF-periodic
features and hard spatial boundary values. For \CDR{}, $\mathcal{A}_f$ is the identity and
initial and boundary conditions enter as loss penalties. These family-specific constructions are detailed in
App.~\ref{app:family}, App.~\ref{app:ccp2}, and App.~\ref{app:fisher}.

\subsection{Training and deployment}
\paragraph{Meta-training.} We sample PDE instances and collocation points and jointly optimize the
hypernetwork, shared bases $C^\ell,R^\ell$, and the backbone's remaining parameters, including its input and
output layers. Gradients pass through the predicted codes and the configured PINN. The objective combines
squared equation residuals with any soft initial/boundary penalties; hard constraints enter through the
coordinate and output transforms above. Each meta-step averages losses over the declared training
structures, then adds the basis penalty once (App.~\ref{app:training}). No solution labels enter this optimization.

\paragraph{Coefficient sampling.} The sampler determines which numerical instances are presented to
the amortizer; the descriptor determines how they are represented. High-reaction \CDR{} uses bounded
perturbations around coefficient-grid nodes, with a radius that gradually expands to cover the coefficient
box (App.~\ref{app:cdr-sampling}). The schedule is shared across descriptors. Appendix~\ref{app:coefficient-sampling}
fixes the seven training structures, objective, and budgets and compares this schedule with full-radius
perturbations from the start and grid-only sampling.

\paragraph{Deployment.} For each target instance, we evaluate $s=H(G)$ once and initialize a separate solver
with the predicted codes and meta-trained backbone. The hypernetwork is no longer optimized. Our main
\emph{full} fine-tune updates the instance's codes and all backbone parameters using the target physics
objective. Fisher--KPP retains a ReLU parameterization of the effective diagonal codes during adaptation;
CDR and CCP optimize the diagonal entries directly (App.~\ref{app:training}).
Numerical reference solutions are used only for evaluation.
Algorithm~\ref{alg:physics-only} in App.~\ref{app:training} summarizes both phases;
App.~\ref{app:protocols} gives the family-specific protocols.

\subsection{Descriptor baselines}
\paragraph{DeepSets-based term set.} We use a standard DeepSets encoder \citep{zaheer2017deep} for
variable-cardinality term conditioning. It independently embeds each term token $z_q$ and mean-pools over the
term set $T$, $r=\rho\big(\tfrac{1}{|T|}\sum_{q\in T}\phi(z_q)\big)$, with its own learned embedding
and readout networks. Tokens contain coefficient sign and log-magnitude, derivative orders, and arity.
In Fisher--KPP, they omit source-field and target-equation assignment and are pooled globally; CCP instead
groups terms by residual/field. Thus the graph--set comparison evaluates complete descriptor designs,
including their input information and readouts.

\paragraph{Coefficient vector.} We adapt the coefficient-conditioned hypernetwork of \citet{cho2023} to
each benchmark family: an MLP maps its predefined coefficient layout to codes. The layouts, normalization,
and missing-term conventions are part of our benchmark implementation. Equation-specific slots can encode
coupling assignments implicitly. Absent terms are zero-filled;
in our standardized \CDR{}
implementation, this zero can coincide with an active coefficient's mean; Fisher--KPP additionally tests
explicit presence indicators. Layouts and transformations are specified in App.~\ref{app:descriptors}.

Each descriptor retains the active numerical coefficients and is trained with its own hypernetwork and
backbone under the same within-family protocol. Backbone architectures and approximate total parameter
counts are matched within each family; the fitted weights are learned separately.

\section{Problem families and evaluation rationale}
\label{sec:data}
We select three PDE families to examine when a hypernetwork benefits from receiving operator composition
and cross-field relationships alongside coefficients. Scalar \CDR{} tests combinations of local operators
without cross-field coupling. Two-field Fisher--KPP adds independently controlled interactions between fields,
allowing familiar couplings to be combined in an unseen pattern. CCP retains a physically meaningful
continuity--Poisson feedback while varying only its coefficients. All three problems have one spatial
dimension, with a separate model trained for each family and descriptor comparisons made within each family.

\subsection{\CDR{}: recombining local operators}
\label{sec:data-cdr}
Convection transports a profile, diffusion smooths it, and logistic reaction drives growth toward saturation.
Their combinations provide a controlled setting for studying equation composition in a single field. Our
\CDR{} family combines the operators studied by
\citet{krishnapriyan2021} and used for parameterized-PDE meta-learning by \citet{cho2023}.
For $u(x,t)$ on the periodic line $x\in[0,1)$, $t\in[0,1]$, the full residual is
\begin{equation}
  \mathcal{R}[u]=\pt{u}\;+\;a\,\px{u}\;-\;D\,\pxx{u}\;-\;r\,u(1-u),
  \label{eq:cdr}
\end{equation}
with three switchable \emph{motifs}: advection, diffusion, and logistic reaction. The evolution term
$\pt{u}$ is always present; each non-empty subset of the three motifs defines one of seven structures
(Table~\ref{tab:structures}). The initial profile, periodic boundary condition, and time window are fixed,
so only the active motifs and their coefficients vary. App.~\ref{app:family} gives the coefficient ranges,
initial profile, and numerical references. Throughout, ``Fisher'' denotes the scalar diffusion--reaction
member; ``two-field Fisher--KPP'' denotes the coupled family below.

\begin{table}[t]
\centering
\caption{The \CDR{} structures and the cross-structure split: train on the five seen structures, test on the two
unseen combinations of advection with reaction.}
\label{tab:structures}
\small
\begin{tabular}{@{}l l ccc@{}}
\toprule
 & residual $=0$ & ADV $a\px{u}$ & DIFF $D\pxx{u}$ & RXN $r\,u(1{-}u)$ \\
\midrule
\multicolumn{5}{@{}l}{\emph{train (5 structures, seen)}}\\
convection & $\pt{u}+a\px{u}$                                   & $\bullet$ &           &           \\
heat       & $\pt{u}-D\pxx{u}$                                  &           & $\bullet$ &           \\
reaction   & $\pt{u}-r\,u(1{-}u)$                               &           &           & $\bullet$ \\
conv-diff  & $\pt{u}+a\px{u}-D\pxx{u}$                          & $\bullet$ & $\bullet$ &           \\
Fisher     & $\pt{u}-D\pxx{u}-r\,u(1{-}u)$                      &           & $\bullet$ & $\bullet$ \\
\addlinespace
\multicolumn{5}{@{}l}{\emph{test (2 structures, unseen combinations)}}\\
adv-reaction      & $\pt{u}+a\px{u}-r\,u(1{-}u)$               & $\bullet$ &           & $\bullet$ \\
adv-diff-reaction & $\pt{u}+a\px{u}-D\pxx{u}-r\,u(1{-}u)$      & $\bullet$ & $\bullet$ & $\bullet$ \\
\bottomrule
\end{tabular}
\end{table}

\paragraph{Term composition and coefficient variation.} The cross-structure protocol trains one amortizer
on five structures and deploys on two unseen combinations. Training never pairs advection with reaction;
both test structures do. This asks whether conditioning on familiar operators helps configure a solver for
their unseen combination.

The reaction coefficient also controls an optimization challenge: PINNs can converge to inaccurate
low-residual solutions at high reaction rates \citep{krishnapriyan2021}. We therefore distinguish
\emph{structural generalization} from \emph{optimization reliability}. The five-to-two protocol tests held
structures in lower-reaction and high-reaction regimes. A separate all-seven-structure protocol trains through the
high-reaction range and evaluates held-out coefficients of known structures; it tests the role of training
coverage and descriptor-conditioned initialization. Section~\ref{sec:exp} compares the high-reaction protocols;
App.~\ref{app:cdr-seeds} reports the lower-reaction results.

\subsection{Two-field Fisher--KPP: recombining cross-field couplings}
\label{sec:data-fisher}
{
Fisher--KPP models a field that spreads by diffusion and grows locally toward saturation
\citep{fisher1937wave}.
\citet{holzer2016proof} studies two such equations with a linear one-way source coupling, showing that
coupling can change spreading dynamics. Our periodic, advective extension controls both coupling
directions independently. On $x\in[0,1)$, $t\in[0,1]$, the two fields satisfy
$\mathcal{R}_u=\mathcal{R}_v=0$, with residuals
\begin{subequations}
\label{eq:fisher}
\begin{align}
  \mathcal{R}_u
  &= \partial_t u-\epsilon_u\partial_{xx}u+a_u\partial_x u
     -\rho_u u(1-u)-\kappa_{uv}v,
  \label{eq:fisher_u}\\
  \mathcal{R}_v
  &= \partial_t v-\epsilon_v\partial_{xx}v+a_v\partial_x v
     -\rho_v v(1-v)-\kappa_{vu}u.
  \label{eq:fisher_v}
\end{align}
\end{subequations}
The source $+\kappa_{uv}v$ in the evolution of $u$ represents $v\to u$, while $+\kappa_{vu}u$ represents
$u\to v$. Setting either nonnegative coupling coefficient to zero removes that interaction while retaining
both fields' transport, diffusion, and growth. The initial fields, periodic boundaries, and time window are
fixed, and the local coefficients and active coupling strengths vary within prescribed ranges
(App.~\ref{app:fisher}). This gives a small coupled family in which source-field and target-equation
assignments can change without changing the local physical mechanisms.

We train on the uncoupled system and each of the two one-way systems, then test the bidirectionally coupled
system at four coefficient settings. Both interactions have appeared during training, but the test activates
them together for the first time. Unlike scalar \CDR{}, this composition specifies not only which operators
are active but also which field influences which equation. The question is whether explicitly organizing these
relationships improves solver conditioning for the unseen coupling combination. App.~\ref{app:fisher}
specifies the held systems; Apps.~\ref{app:fisher-evidence} and~\ref{app:fisher-ownership}
report the descriptor controls.
}

\subsection{CCP: coefficient variation within a fixed coupled structure}
\label{sec:data-ccp2}

A capacitively coupled plasma (CCP) is a gas discharge driven by an alternating voltage between electrodes,
with charge transport and the electric field influencing each other.
This fixed-coupling family tests whether physical interaction alone motivates graph conditioning.
We use a simplified one-dimensional drift--diffusion--Poisson model \citep{jing2026extending},
coupling a charged-species density $N$ to an electrostatic potential $\Phi$.
On $\xi,\tau\in[0,1]$, its residuals are
\begin{subequations}
\label{eq:ccp2} %
\begin{align}
  \mathcal{R}_N &= \partial_\tau N-a_D\,\partial_{\xi\xi}N+a_\mu\,\partial_\xi\!\big(N\,\partial_\xi\Phi\big)-S(\xi), \label{eq:ccp2_N} \\ %
  \mathcal{R}_\Phi &= \partial_{\xi\xi}\Phi-\beta\,(N-1), \label{eq:ccp2_Phi} %
\end{align}
\end{subequations}
The potential gradient drives density transport in $\mathcal{R}_N$, and the density sources the potential in
$\mathcal{R}_\Phi$. Both feedback links are retained throughout the family: deleting either would remove
part of this self-consistent coupling. Instead, an applied voltage $V_0$ and source rate $R_0$ determine the
varying coefficients while the operator structure stays fixed. We train on a continuous $(V_0,R_0)$ box
and deploy on held-out interior parameter values. The nondimensional map, boundary conditions, and
reference fields are given in App.~\ref{app:ccp2}.

A fixed coefficient layout describes the instance variation in this family. CCP therefore tests whether
graph-based processing provides an additional benefit in a coupled application whose connectivity is already
known, complementing Fisher--KPP's varying-coupling test.

Our working hypothesis across these problems is that the utility of relational conditioning depends on which
parts of an equation vary across instances, rather than on the presence of coupling alone. Section~\ref{sec:exp}
evaluates the three roles through within-family comparisons of accuracy after target-specific adaptation.

\section{Experiments}
\label{sec:exp}
\subsection{Evaluation protocol}
\label{sec:exp-protocol}
We evaluate equation-conditioned initializations by the solution accuracy reached within a fixed budget
of target-specific physics optimization.
We compare coefficient-vector, term-set, and operator-graph conditioning using the shared factorized
backbone of \S\ref{sec:method}. Each family has its own trained models. Within a comparison, the
amortized descriptor methods share the PDE distribution, backbone architecture, physics objective, sampling policy,
and training and adaptation budgets. The from-scratch control uses the same backbone and deployment
budget without meta-training. Main results use ten training seeds for \CDR{} and five for each coupled
family; their numerical settings are in Apps.~\ref{app:cdr-ten-seed},~\ref{app:fisher}, and~\ref{app:ccp2}.

We measure relative solution error
$\rell=\lVert u_\theta-u^\star\rVert_2/\lVert u^\star\rVert_2$ on the reference grid, averaging
fieldwise errors for coupled systems. We first average held cases within each training seed, then report
the mean and sample standard deviation across independent seeds. For Fisher--KPP and CCP, the primary
measure is the \emph{final} error after 3,000 adaptation steps. The \CDR{} optimization study reports
\emph{oracle-best}: the lowest error over scheduled evaluations within its fixed deployment budget.
Reference solutions are used only for evaluation, including oracle-best selection, and never for training
or online stopping. Results from distinct training protocols are reported separately.

\subsection{Term composition and training coverage in \CDR{}}
\label{sec:exp-cdr}
Scalar \CDR{} first tests whether term-based conditioning helps when familiar operators appear in new
combinations. The high-reaction five-to-two and all-seven protocols of
\S\ref{sec:data-cdr} evaluate the same eight high-reaction AR/ADR cases, separating unseen term composition
from held-coefficient adaptation within known structures.

\begin{table}[!htbp]
\centering
\caption{High-reaction \CDR{} oracle-best $\rell$ (mean $\pm$ sample SD over ten seed means).
Both protocols evaluate the same eight AR/ADR cases. Five-to-two holds out their structures;
all-seven training holds out only their coefficients.}
\label{tab:cdr-ten-seed}
\small
\setlength{\tabcolsep}{5pt}
\begin{tabular}{@{}lccc@{}}
\toprule
Meta-training & Coefficient vector & Term set & Operator graph \\
\midrule
Five-to-two & $0.7459\pm0.3690$ & $0.1112\pm0.0979$ & $0.1093\pm0.1221$ \\
All seven   & $0.4869\pm0.4803$ & $0.0191\pm0.0016$ & $0.0204\pm0.0018$ \\
\bottomrule
\end{tabular}
\end{table}

Table~\ref{tab:cdr-ten-seed} shows substantially lower mean errors for both term-based descriptors
than for the coefficient vector. On unseen structures, the term set and operator graph reduce the mean
by $85.1\%$ and $85.4\%$, respectively. Including all seven structures further lowers their errors and
reduces seed-to-seed variation. Their means remain close in both protocols: this scalar family supports
term-based conditioning, without a consistent advantage for the graph over the set. Fixed-final errors
support the same comparison (App.~\ref{app:cdr-seeds}), which also reports lower-reaction results and seed distributions.

A supplementary all-seven experiment (App.~\ref{app:coefficient-sampling}) tests dependence on the
coefficient-sampling schedule. Low Set and Graph errors persist with full jitter from the
start. Under grid-only training, Graph has more consistent threshold attainment across the sampled
seeds. Thus coefficient exposure affects adaptation alongside the descriptor; these supplementary
results use their own collocation budget and seed cohorts.

\subsection{Generalization to unseen Fisher--KPP coupling}
\label{sec:exp-fisher}
\begin{wrapfigure}{r}{0.46\textwidth}
\vspace{-8pt}
\centering
\captionsetup{skip=3pt,justification=raggedright,singlelinecheck=false}
\includegraphics[width=\linewidth]{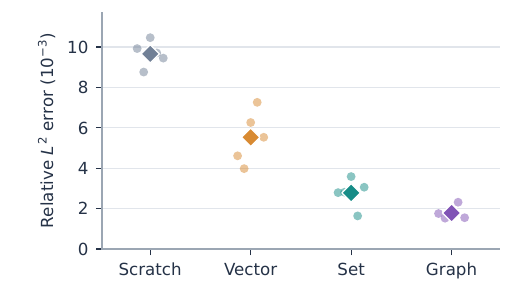}
\caption{Fisher--KPP error at step 3000. Circles: five seed means over H1--H4; diamonds: overall means.}
\label{fig:fisher-final}
\end{wrapfigure}
Fisher--KPP tests whether explicitly representing cross-field relationships improves adaptation to an
unseen combination of familiar interactions. Under the split in \S\ref{sec:data-fisher}, all four held
systems activate both coupling directions, which meta-training sees only separately.

The operator graph has the lowest mean final error (Fig.~\ref{fig:fisher-final}):
$(1.786\pm0.318)\times10^{-3}$, compared with $(2.779\pm0.712)\times10^{-3}$ for the term set,
$(5.531\pm1.302)\times10^{-3}$ for the coefficient vector, and
$(9.660\pm0.626)\times10^{-3}$ from scratch. Its mean is $35.7\%$ below the term set and
$67.7\%$ below the coefficient vector. The ordering is also preserved under oracle-best evaluation
(Table~\ref{tab:fisher-aggregate}). The lower final error supports graph conditioning for adaptation to the
unseen coupling combination, beyond the gains already provided by the term set.
The lower Graph mean also holds for each held case and each field separately
(App.~\ref{app:fisher-evidence}, Table~\ref{tab:fisher-per-case}).

\begin{figure}[!htbp]
\centering
\includegraphics[width=0.85\linewidth]{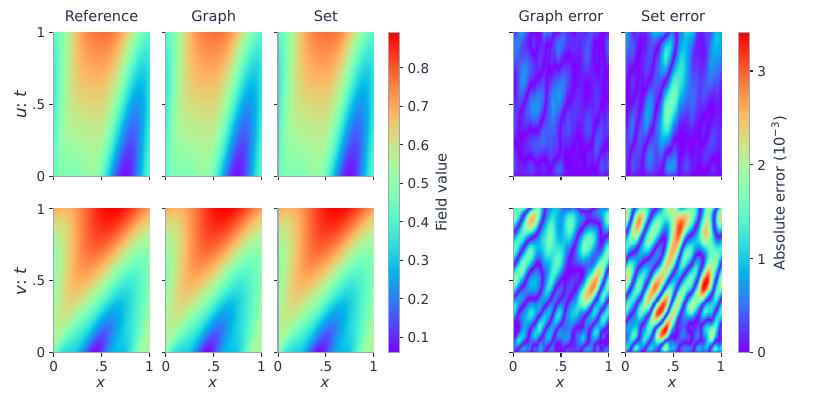}
\vspace{-5pt}
\caption{Fisher--KPP H1, seed 101. Rows show $u$ and $v$; columns show the reference, Graph and Set
predictions, and their absolute errors. The first three columns share the adjacent field-value colorbar;
the error columns share the rightmost linear colorbar. Both use rainbow. The case and seed are fixed;
Graph and Set use their respective oracle-best checkpoints at steps 2400 and 3000. These snapshots
illustrate solution quality; Fig.~\ref{fig:fisher-final} reports aggregate final-step performance.}
\label{fig:fisherfields}
\end{figure}

The solution maps in Fig.~\ref{fig:fisherfields} show that both learned solvers recover the main field
structure. Their differences are clearer in the shared-scale absolute-error maps, where the graph
prediction has smaller errors on this example. The quantitative conclusion above uses all held cases
and seeds rather than the displayed snapshot.

A separate five-seed complete-method comparison on one GPU (App.~\ref{app:fisher-iwata}) evaluates
a parameter-matched Iwata-style conditional decoder. It has lower initialization error, whereas Graph
has $44.6\%$ lower mean error after 3,000 updates.

\Needspace{15\baselineskip}
\subsection{Coefficient variation within fixed CCP coupling}
\label{sec:exp-ccp}
\begin{wraptable}{r}{0.44\textwidth}
\vspace{-20pt}
\centering
\captionsetup{skip=3pt,justification=raggedright,singlelinecheck=false}
\captionsetup{position=top}
\caption{CCP error at step 3000: mean $\pm$ sample SD over five seeds.}
\label{tab:ccp-final}
\small
\begin{tabular}{@{}lc@{}}
\toprule
Method & Final $\rell$ \\
\midrule
From scratch       & $0.4866\pm0.0435$ \\
Coefficient vector & $0.1396\pm0.0347$ \\
Term set           & $0.2361\pm0.0782$ \\
Operator graph     & $0.2483\pm0.1152$ \\
\bottomrule
\end{tabular}
\end{wraptable}
CCP retains the continuity--Poisson feedback throughout training and testing, so coefficients specify
all variation within the equation family. Table~\ref{tab:ccp-final} reports final accuracy at the two
held interior parameter settings. The coefficient vector performs best. The operator graph lowers the
mean error relative to training from scratch, but not relative to coefficient conditioning. This
fixed-structure application shows that physical coupling alone does not ensure a graph advantage under
this protocol.
\WFclear

\subsection{Number of message-passing rounds}
\label{sec:exp-depth}
\begin{wrapfigure}{r}{0.5\textwidth}
\vspace{-8pt}
\centering
\captionsetup{skip=3pt,justification=raggedright,singlelinecheck=false}
\includegraphics[width=\linewidth]{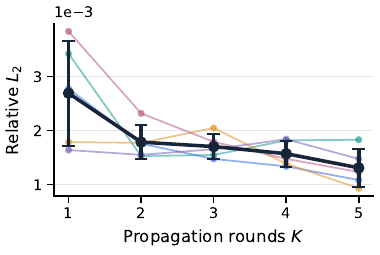}
\caption{Fisher--KPP error after 3000 updates. Thin lines: five seed means over fields and
H1--H4; bold line and whiskers: mean and sample SD.}
\label{fig:fisher-depth-final}
\end{wrapfigure}
On Fisher--KPP, we vary propagation rounds $K\in\{1,2,3,4,5\}$, matching meta-training and
target-encoding depths. Shared weights keep all models at 132,866 parameters;
initialization and coefficient/collocation streams are paired within each seed.

A five-seed depth ablation (App.~\ref{app:fisher-depth}) fixes the parameter count and optimizer
budgets. Increasing propagation from two to five rounds further lowers mean final error, with a
paired improvement in four of the five seeds. The descriptor comparison above retains two rounds.
At step 3000, the mean error reduction from $K=2$ to $K=5$ is $26.6\%$
(Fig.~\ref{fig:fisher-depth-final}).
The comparison fixes optimizer updates, not compute time; deeper encoding requires more computation.
\WFclear

\paragraph{Evidence across families.} Fisher--KPP provides the direct evidence that graph conditioning
improves adaptation to an unseen coupling composition. High-reaction \CDR{} shows benefits from both
term-based descriptors, while CCP identifies a fixed-structure setting where coefficient conditioning performs best.
Together, these observations motivate testing relational conditioning when cross-field interactions vary
across instances. The sampler controls in App.~\ref{app:coefficient-sampling} complement this structural
comparison by testing sensitivity to numerical coefficient exposure.
Complementing fixed-budget accuracy, a separate three-seed \CDR{} study in
App.~\ref{app:amortization} measures updates to a common error threshold. On held structures,
graph conditioning reduces the median from 2200 scratch updates to 600; Vector and Set also
require fewer updates, establishing an amortization benefit shared by the descriptors.

\section{Related work}
\label{sec:related}
\paragraph{Amortized physics-informed solvers.} HyperPINN \citep{de2021hyperpinn} predicts PINN weights
from equation parameters to share learning across instances. Hyper-LR-PINN \citep{cho2023} predicts
diagonal codes for shared factorized layers, including for a parameterized CDR family. We retain this
factorization and coefficient-to-code interface while broadening the equation descriptor. Our comparisons
use full adaptation of codes and backbone parameters (\S\ref{sec:method}), whereas Cho et al.'s online
phase freezes the factorized bases. The contribution concerns solver conditioning within this framework.
In a related dynamics-learning setting, \citet{jing2026structure} use latent-conditioned hypernetworks
for few-shot adaptation of structure-preserving models from trajectory data, whereas we condition PINNs
on explicit equation graphs and adapt them using physics residuals.

\paragraph{Representing equation families.} Coefficient dictionaries can describe different equations,
not just different parameter values of one equation. \citet{iwata2023meta} encode coefficients of a
polynomial in fields and derivatives, combine them with a set representation of boundary conditions,
and condition a coordinate-based solution network trained with physics losses.
Their target-specific refinement updates the solution decoder and the inferred problem representation
while keeping the problem encoder fixed. This differs from coefficient-to-code hypernetwork conditioning.
Appendix~\ref{app:fisher-iwata} compares a parameter-matched two-field adaptation of this conditional
decoder with our solver. Their setting motivates
comparing equation representations without treating fixed-length vectors as intrinsically restricted to
one structure. Our operator graph makes factor and residual assignments explicit. The term-set baseline
uses standard DeepSets pooling \citep{zaheer2017deep}, providing a variable-size representation of additive
terms against which to assess the complete graph descriptor.

\Needspace{5\baselineskip}
\paragraph{Graph-conditioned neural fields.} Graph hypernetworks predict weights from target-network
connectivity in architecture search, unseen-architecture prediction, and program synthesis
\citep{zhang2018graph,knyazev2021parameter,li2026structure}. Our graph instead describes the governing
equations. More closely, PDEformer and PDEformer-2 combine equation graphs with graph Transformers and
implicit neural representations \citep{ye2024pdeformer,pdeformer2}. PDEformer generates layer modulations
from graph embeddings, so graph-conditioned neural fields already have a direct precedent. These models
use numerical solution datasets for pretraining; we use equation graphs to configure factorized PINNs
through physics-only meta-training and adaptation.

\paragraph{Multi-equation learning with physics.} HyPINO \citep{bischof2026hypino} generates PINNs
from PDE parameterizations using manufactured-solution labels and physics losses, and evaluates both
forward prediction and subsequent fine-tuning. PI-MFM \citep{zhu2025pi} combines symbolic equations and
condition inputs, assembles physics losses automatically, and supports physics-only adaptation after
mixed data/physics pretraining. Our study shares the goal of reusing solvers across equations, but focuses
on how coefficient, term-set, and graph descriptors condition a common within-family backbone when
neither meta-training nor adaptation uses solution labels.

\section{Limitations and future directions}
\label{sec:limits}
Our results concern three separately trained, one-spatial-dimensional PDE families with fixed conditions
and bounded coefficients. Transfer across families, dimensions, and boundary conditions remains open.
Varying coefficients and admissible connectivity independently within a common coupled model would
better isolate when relational conditioning helps.

The sampler controls use small, unequal seed cohorts (App.~\ref{app:coefficient-sampling}).
The update-count study (App.~\ref{app:amortization}) and measured Fisher timings
(App.~\ref{app:fisher-iwata}) cover fixed deployment sets. Broader deployment coverage and
physics-based stopping rules remain future work.

\clearpage
\section*{Acknowledgements}
Cheng Jing acknowledges funding support from Applied Materials, Inc. Yixuan He acknowledges support from a Jetstream2 AI Fellowship and an NVIDIA Academic Grant Award.

\bibliographystyle{unsrtnat}
\bibliography{\LoGAssetRoot reference_revision}

\clearpage
\appendix
\raggedbottom
\section{Representation and implementation}
\label{app:graph}
This section gives the descriptor details supplementing \S\ref{sec:method}.
Benchmark settings are in App.~\ref{app:protocols}; supplementary descriptor comparisons
and coefficient-sampling controls are in Apps.~\ref{app:comparisons} and~\ref{app:coefficient-sampling}.

\subsection{Graph construction}
We use $u_f$ for field $f$, $\mathcal R_f$ for its governing residual, $G=(V,E)$ for
the operator graph, and $s_f^\ell$ for the code of factorized layer $\ell$.
Each additive monomial is a term node with a scalar coefficient and a tuple of field
or derivative factors. Arity counts factors with multiplicity: $u^2$ has arity two,
a constant has arity zero, and $-\kappa v$ in $\mathcal R_u$ has arity one with source
field $v$ and target equation $u$.

The four node kinds are fields, derivatives, terms, and residual roots.
Connections run from a field to its derivatives, from factors to their terms,
and from each term to its target residual. Node attributes identify kind, field or
target-field identity, derivative orders, and coefficient sign/log-magnitude where applicable.
The symbolic incidence is directed; message passing uses symmetrized, self-looped,
row-normalized connectivity without edge-specific transformations.
The CDR and CCP layouts reserve a global-node kind but remove that node before encoding.

\begin{figure}[htbp]
\centering
\includegraphics[width=0.75\linewidth]{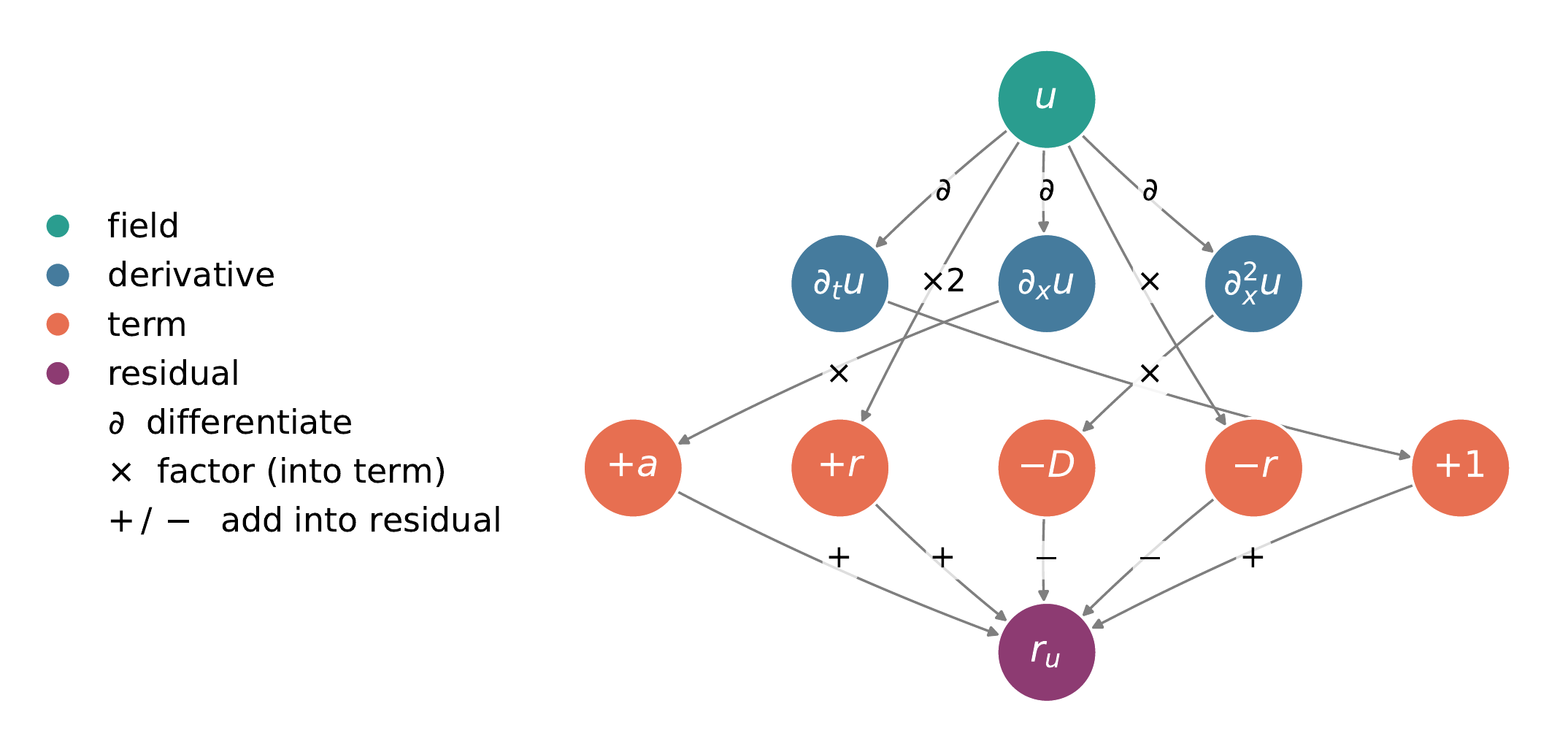}
\caption{CDR operator graph. Other motif combinations remove the corresponding terms and incidences.}
\label{fig:graph-adr}
\end{figure}

\begin{figure}[htbp]
\centering
\includegraphics[width=0.82\linewidth]{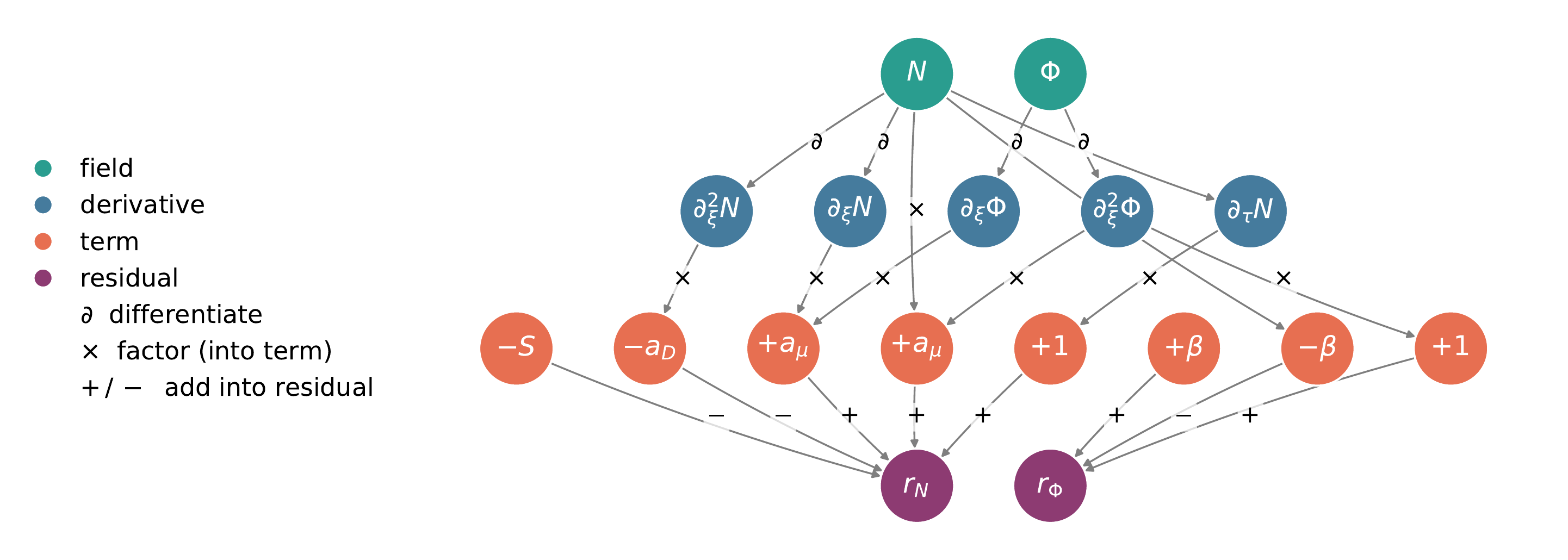}
\caption{CCP operator graph. Density and potential enter both the drift--diffusion and Poisson residuals.}
\label{fig:graph-ccp2}
\end{figure}

Figure~\ref{fig:graph-adr} shows all CDR motifs. Removing reaction removes the $-ru$
and $+ru^2$ terms and their incidences. In a coupled system, the target connection
also identifies which equation receives each term (Fig.~\ref{fig:graph-ccp2}).
The schematics use $r_u,r_N,r_\Phi$ for $\mathcal R_u,\mathcal R_N,\mathcal R_\Phi$.
Their edge signs illustrate residual assembly; the encoders receive coefficient signs
as term-node attributes and use unsigned message-passing adjacency.

\subsection{Family-specific descriptor inputs and readouts}
\label{app:descriptors}
\paragraph{CDR.}
\label{app:cdr-descriptors}
The coefficient vector has layout $[a,D,r]$. Active $D$ uses $\log_{10}$; $a$ and $r$
use the identity transform. Active entries are standardized using the protocol's active-value
means and standard deviations; absent entries are zero-filled after standardization, without
presence indicators. Thus absence and a mean-valued active coefficient can share an encoding.
Term-set and graph inputs instead use coefficient sign and log-magnitude without this standardization.

The graph has 22-dimensional features: five node-kind slots, eight field-identity slots,
three derivative-order slots, three term-attribute slots, and three source/location slots.
Its two residual update rounds have round-specific affine maps. The embedded and updated
states are concatenated at each node, projected to the node width, and mean-pooled over
term and field nodes before decoding the single field's codes. Each amortized CDR model
has approximately $1.0$M parameters.

\paragraph{Fisher--KPP.}
\label{app:fisher-descriptors}
The equation-ordered vector is
$[\epsilon_u,a_u,\rho_u,\kappa_{uv},\epsilon_v,a_v,\rho_v,\kappa_{vu}]$.
Diffusion entries are $\log_{10}$-transformed, active entries are analytically standardized,
and absent couplings are filled with zero after standardization.
The presence-indicator control appends eight binary indicators.

A term-set token contains
$[\mathrm{sign}(c),\log_{10}|c|,\sum d_x,\sum d_t,\mathrm{arity},
\mathbb 1_{\mathrm{constant\ source}}]$.
Tokens are independently embedded and globally mean-pooled; the readout then receives
target-field identity. The tokens do not identify their source fields or target equations.
The ownership-aware control adds these assignments (App.~\ref{app:fisher-ownership}).

The graph has 12-dimensional node features: four node-kind indicators, two field indicators,
two derivative orders, and coefficient sign, log-magnitude, arity, and a constant-source flag.
The bidirectionally coupled test graph has 22 nodes and 30 directed incidence edges before
symmetrization. Its update and readout follow \S\ref{sec:method}.
The operator graph uses two rounds and has 132,866 parameters. Other descriptor totals
are within $0.3\%$ of this total.

\paragraph{CCP.}
\label{app:ccp-descriptors}
The vector uses a 17-slot union layout with four active coefficients
$(a_D,a_\mu,\beta,S_0)$, where $S_0$ is the source amplitude defined in App.~\ref{app:ccp2}.
Active entries use $\log_{10}(|c|+10^{-8})$ without standardization; other slots are zero.
One predicted code set is broadcast to both fields. The set encoder uses five term attributes
(coefficient sign, log-magnitude, two derivative orders, and arity) and groups terms by residual/field.
The graph uses the 22-dimensional layout above, including source-location attributes, with
two non-residual update rounds and a final-hop per-field readout. Total parameter counts
differ by at most $0.12\%$ between amortized arms.

\subsection{Consistency between the graph and physics loss}
CDR and Fisher--KPP share term lists between descriptor construction and automatic-differentiation
residual evaluation. The five-seed CCP implementation constructs the graph explicitly while
differentiating its drift in conservative form:
\[
 \partial_\xi(N\,\partial_\xi\Phi)
 =(\partial_\xi N)(\partial_\xi\Phi)+N\,\partial_{\xi\xi}\Phi.
\]
Both right-hand terms occur in the graph with coefficient $a_\mu$.
The Poisson residual is $\partial_{\xi\xi}\Phi-\beta(N-1)$.
The graph and physics loss use the same coefficients and prescribed two-zone source:
the graph stores the first interval and uses the fixed mirrored-interval convention.
Thus the two code paths represent the same governing equations.

\paragraph{Engineering value.}
Within the supported operator vocabulary, the graph provides an inspectable interface
between an equation definition and the hypernetwork. In CDR and Fisher--KPP, one term
representation supplies both graph construction and residual evaluation, so supported
term or coupling edits can be expressed once and reused in both paths. This code reuse
complements the accuracy comparisons; new operators, field layouts, or boundary
prescriptions still require explicit implementation.

\FloatBarrier
\subsection{Physics-only training and adaptation}
\label{app:training}
Algorithm~\ref{alg:physics-only} summarizes the two phases for one PDE family.
The physics loss is the weighted mean of squared equation residuals, computed by
automatic differentiation, plus any soft initial/boundary penalties. Hard conditions
are built into the coordinate and output transforms. Numerical reference solutions
enter only the separate evaluation procedure (App.~\ref{app:metrics}).

\begin{algorithm}[!htbp]
\caption{Physics-only meta-training and full adaptation.}
\label{alg:physics-only}
\begin{algorithmic}[1]
\Statex \textbf{Input:} Training PDE structures, coefficient sampler, prescribed conditions,
and training/adaptation budgets.
\Statex \textbf{Phase 1: Learn a shared equation-conditioned initialization}
\State Initialize the graph hypernetwork and shared factorized PINN backbone.
\For{each meta-training step}
  \State Sample one PDE per training structure and fresh collocation points.
  \For{each sampled PDE}
    \State Build its operator graph, retaining coefficients and term assignments.
    \State Predict diagonal codes with the hypernetwork and configure the shared PINN.
    \State Evaluate the physics loss for this PDE at its sampled points.
  \EndFor
  \State Average the PDE losses; add the configured basis-orthogonality penalty once.
  \State Backpropagate through the codes; update the hypernetwork and shared backbone.
\EndFor
\Statex \textbf{Phase 2: Adapt an independent solver to each target equation}
\For{each target PDE}
  \State Build its graph and call the trained hypernetwork once to obtain initial codes.
  \State Copy the trained backbone; make the codes independent trainable tensors.
  \For{each adaptation step}
    \State Sample fresh points and recompute the target physics loss and configured penalty.
    \State Update the codes and all copied backbone parameters; keep the hypernetwork fixed.
  \EndFor
  \State Return the adapted PINN as the target solution approximation.
\EndFor
\end{algorithmic}
\end{algorithm}

Meta-training is joint optimization across equations; it has no inner adaptation
loop. The training set has five or seven structures in CDR, three in Fisher--KPP,
and one in CCP. Equal step counts across different structure sets therefore need
not imply equal computation. Samplers, loss weights, regularization, and optimizer
budgets follow the family protocols in App.~\ref{app:protocols}.

The predicted codes include the output ReLU of Eq.~\eqref{eq:codes}.
At deployment, Fisher--KPP initializes independent trainable tensors
$\tilde s_f^\ell$ with these predictions and uses
$s_f^\ell=\operatorname{ReLU}(\tilde s_f^\ell)$ in
$C^\ell\operatorname{diag}(s_f^\ell)R^\ell$ at every forward pass.
CDR and CCP instead optimize $s_f^\ell$ directly, without a nonnegativity clamp.
Full adaptation updates the instance's code tensors and all copied backbone
parameters; the hypernetwork remains unchanged.

\FloatBarrier
\section{Benchmark and evaluation protocols}
\label{app:protocols}
\FloatBarrier
\subsection{Errors and aggregation}
\label{app:metrics}
For one field, $\rell=\|u_\theta-u^\star\|_2/\|u^\star\|_2$ on the reference grid.
For a coupled case, we average the two fieldwise relative errors.
Final error is measured at the fixed deployment budget; oracle-best is the minimum
case error over scheduled evaluations, including initialization. Reference fields
are used only for evaluation, not for training or online stopping.

We average case errors within each training seed, then report the mean and sample
standard deviation over ten CDR or five Fisher--KPP/CCP seed means.
The three-seed ownership control and the supplementary CDR sampling cohorts are
reported separately in Apps.~\ref{app:fisher-ownership} and~\ref{app:coefficient-sampling}.

\FloatBarrier
\subsection{CDR: motif combinations and coefficient regimes}
\label{app:family}
The scalar family of Eq.~\eqref{eq:cdr} combines convection, diffusion, and logistic reaction
\citep{krishnapriyan2021,cho2023} on $x\in[0,1)$ and $t\in[0,1]$.
Advection translates the profile, diffusion damps spatial variation, and logistic reaction
drives positive values toward saturation. The seven nonempty combinations of these motifs
are listed in Table~\ref{tab:structures}. All share the initial condition
\[
 u_0(x)=0.5+0.20\sin(2\pi x)+0.10\sin(4\pi x+0.7)+0.05\cos(6\pi x)
\]
and periodic boundary conditions. Both constraints enter as separate unit-weighted soft penalties.
In the five-to-two split, meta-training omits AR and ADR; these two structures combine
advection and reaction for the first time at deployment. In the all-seven protocol,
all structures are seen and only evaluation coefficient settings are new.

\paragraph{Reference solutions.}
We use 256 endpoint-excluded Fourier spatial nodes and 101 output times, with a float64
pseudo-spectral method of lines, BDF integration, relative tolerance $10^{-8}$, and absolute
tolerance $10^{-10}$, without dealiasing. Translation and diffusion modal-decay checks give
errors of approximately $10^{-7}$. References are used only for evaluation.

\FloatBarrier
\subsubsection{Ten-seed descriptor comparisons}
\label{app:cdr-ten-seed}
The ten-seed comparisons use the width-256 backbone with three coded layers and two graph-update
rounds. Each physics-loss evaluation draws 512 interior, 128 initial, and 128 periodic-boundary
samples. Meta-training and full adaptation use Adam at $2.5\times10^{-4}$; the basis-orthogonality
weight is $10^{-4}$ during meta-training and zero during adaptation. Errors are evaluated every 200 deployment steps. All three protocols use seeds
$0$--$9$ and float32 computation on NVIDIA GB10 GPUs. Task draws share seed IDs across descriptors;
collocation sampling follows the same policy but is not pointwise paired across architectures.

Both high-reaction protocols use 30,000 meta-training and 3,000 deployment steps. They share the
coefficient grid and bounded-jitter schedule of App.~\ref{app:cdr-sampling}, with radius held at zero
through step 8,000 and expanded to full coverage by step 24,000. Each meta-step averages the loss
over the declared five or seven training structures. The five-to-two protocol omits AR/ADR losses;
the all-seven protocol includes them. Their common primary evaluation comprises AR and ADR at
$a\in\{4.5,9.5\}$ and $r\in\{8.5,9.5\}$, with $D=10^{-2}$ for ADR, giving eight cases.
Table~\ref{tab:cdr-ten-seed} reports their results.

The lower-reaction five-to-two protocol instead uses 20,000 meta-training and 5,000 deployment
steps, with $a\sim U[2,4]$, $r\sim U[1,4]$, and log-uniform $D\in[10^{-3},10^{-1}]$ on active
terms. App.~\ref{app:cdr-seeds} reports its six-case held AR/ADR evaluation.

\FloatBarrier
\subsubsection{Coefficient grid and sampling schedule}
\label{app:cdr-sampling}
The high-reaction protocols sample the Cartesian product of the active coefficient axes
in Table~\ref{tab:grid}. Write $\alpha(m)\in[0,1]$ for the jitter radius at meta-step $m$.
A sampled grid node is perturbed uniformly by up to $\alpha(m)$ times the axis half-spacing,
in $\log_{10}D$ for diffusion, then clipped to its range. At $\alpha=0$ only grid nodes are
sampled; at $\alpha=1$ neighboring jitter intervals meet and the whole box is covered.
This is continuous support, not a claim that the clipped distribution is uniform.
The scheduled setting holds $\alpha=0$ through step 8,000 and increases it linearly to one
by step 24,000 of 30,000.
\begin{table}[!htbp]
\centering
\caption{High-reaction coefficient grid and jitter half-spacings.}
\label{tab:grid}
\small
\begin{tabular}{@{}l l l l@{}}
\toprule
axis & nodes & spacing (HALF) & range \\
\midrule
$a$ (advection) & $2,\ 4.5,\ 7,\ 9.5,\ 12$ (linear)                    & $2.5$ ($1.25$)   & $[2,\ 12]$ \\
$D$ (diffusion) & $10^{-3},10^{-2.5},10^{-2},10^{-1.5},10^{-1}$ (log)  & $0.5$ dec ($0.25$)& $[10^{-3},\ 10^{-1}]$ \\
$r$ (reaction)  & $1,\ 3.25,\ 5.5,\ 7.75,\ 10$ (linear)               & $2.25$ ($1.125$) & $[1,\ 10]$ \\
\bottomrule
\end{tabular}
\end{table}

\FloatBarrier
\subsection{Two-field Fisher--KPP: held coupling composition}
\label{app:fisher}
We solve the residual system in Eqs.~\eqref{eq:fisher_u}--\eqref{eq:fisher_v} on $x\in[0,1)$,
$t\in[0,1]$, with periodic space and
\begin{align*}
u_0(x)&=0.35+0.20\sin(2\pi x)+0.08\cos(4\pi x+0.3),\\
v_0(x)&=0.30+0.18\cos(2\pi x+0.4)-0.06\sin(6\pi x).
\end{align*}
For each $q\in\{u,v\}$, we independently draw $\log_{10}\epsilon_q$ uniformly on
$\log_{10}[0.005,0.02]$, $a_q\sim U[0.2,0.8]$, and $\rho_q\sim U[0.8,1.6]$; each active coupling is drawn
from $U[0.1,0.4]$. The three meta-training structures are uncoupled
($\kappa_{uv}=\kappa_{vu}=0$), $v\to u$ only ($\kappa_{uv}>0,\kappa_{vu}=0$), and $u\to v$ only
($\kappa_{uv}=0,\kappa_{vu}>0$). Within a meta-training step, they share one draw of the six local coefficients
and the same collocation batch. The four test systems activate both couplings. The residual convention makes
the advection term positive on the left-hand side and the active coupling term negative; the equivalent
evolution equations have $-a_q\partial_xq$ and positive cross-field sources.

\begin{table}[htbp]
\centering
\caption{Four held bidirectionally coupled Fisher--KPP coefficient settings.}
\label{tab:fisher-held}
{
\scriptsize
\begin{tabular*}{\linewidth}{@{\extracolsep{\fill}}lrrrrrrrr@{}}
\toprule
task & $\epsilon_u$ & $a_u$ & $\rho_u$ & $\kappa_{uv}$ & $\epsilon_v$ & $a_v$ & $\rho_v$ & $\kappa_{vu}$\\
\midrule
H1&.008&.25&.90&.15&.018&.65&1.50&.35\\
H2&.015&.55&1.40&.32&.006&.30&1.00&.18\\
H3&.010&.75&1.20&.25&.012&.45&1.30&.40\\
H4&.017&.40&1.55&.38&.009&.70&.85&.22\\
\bottomrule
\end{tabular*}
}
\end{table}
The backbone has width 96, three coded factor layers, \texttt{tanh} activation, orthogonal basis initialization,
and spatial Fourier features at frequencies $1,2,3,4$. The initial condition is imposed exactly as
$(u,v)=(u_0,v_0)+t f_\theta(x,t)$. Training seeds are $101,202,303,404,505$. Meta-training uses 3,000
Adam steps at $5\times10^{-4}$; deployment uses 3,000 steps at $2\times10^{-4}$. Both phases resample 1,024
interior points per step, clip the gradient norm at 10, weight the two residuals equally, and set the basis penalty
to zero. Scratch uses the same width-96 backbone without an encoder and is optimized under the same deployment
objective, test settings, sampled batches, and 3,000-step budget. We evaluate every 100 steps. Descriptor configurations are specified in App.~\ref{app:fisher-descriptors}.

Reference fields use a float64 periodic Fourier method of lines on a 256-point spatial grid with 101 uniform output
times, integrated by BDF with relative tolerance $10^{-9}$, absolute tolerance $10^{-11}$, and maximum step
$0.02$. The relative difference between reference fields computed on 128- and 256-point grids is less than
$10^{-12}$. References are used only for evaluation.

\FloatBarrier
\subsection{CCP: fixed-structure coefficient variation}
\label{app:ccp2}
The one-dimensional continuity--Poisson model of Eq.~\eqref{eq:ccp2}
\citep{jing2026extending} evolves density $N$ and potential $\Phi$ on
$\xi\in[0,1]$, with RF phase $\tau\in[0,1]$. Potential drives density transport,
and density supplies the Poisson source. The structure is fixed throughout this benchmark.

\paragraph{Nondimensionalization and constraints.}
With $n_{\mathrm{io}}=R_0(x_2-x_1)/c_s$,
\[
 a_D=\frac{DT}{L^2},\qquad a_\mu=\frac{\mu V_0T}{L^2},\qquad
 \beta=\frac{e\,n_{\mathrm{io}}L^2}{\varepsilon_0 V_0},\qquad
 S(\xi)=S_0 I(\xi),\quad S_0=\frac{R_0T}{n_{\mathrm{io}}},
\]
where $I$ is the indicator of $[0.2,0.4]\cup[0.6,0.8]$ and $x_2-x_1=0.2L$.
Thus $\beta\propto R_0/V_0$, while $S_0=Tc_s/(x_2-x_1)$ is independent of $R_0$.
The physical constants are $L=0.025$ m, $f=13.56$ MHz, $T=1/f$, $T_e=3.0$ eV,
$\nu_m=10^8\,\mathrm{s}^{-1}$, and $m_i=40$ amu, with
$\mu=e/(m_e\nu_m)$, $D=\mu T_e$, and $c_s=\sqrt{eT_e/m_i}$.
There is no initial-condition constraint. Spatial boundary values are imposed through
\[
 N=\xi(1-\xi)e^{g_N},\qquad
 \Phi=\xi\sin(2\pi\tau)+\xi(1-\xi)g_\Phi .
\]
\paragraph{Reference bank and evaluation cases.}
The reference bank covers $V_0\in\{130,200,270\}$ V and
$R_0\in\{1,2,3.5,5\}\times10^{20}\,\mathrm{m}^{-3}\mathrm{s}^{-1}$.
It uses 200 finite-volume cells (201 nodes), backward Euler with Scharfetter--Gummel
fluxes, 2,000 time steps per RF period, three Gummel sweeps per step, and 40 periods,
storing 200 time levels from the final period. The bank filter rejects nonfinite values,
a mid-domain bulk mean $N<0.7$, or $\max|\Phi|>3$.
We evaluate the two interior cells $V_0=200$ V,
$R_0\in\{2,3.5\}\times10^{20}\,\mathrm{m}^{-3}\mathrm{s}^{-1}$.
Training samples the continuous enclosing box without excluding these neighborhoods,
so this is coefficient interpolation, not a held region or held topology.

\FloatBarrier
\subsubsection{Five-seed training and deployment}
\label{app:ccp-primary-protocol}
The width-96 backbone has three coded layers and coordinate input
$[\xi,\sin(2\pi\tau),\cos(2\pi\tau)]$, enforcing RF periodicity exactly.
Seeds 0--4 each receive 3,000 meta-training steps and 3,000 full-adaptation steps;
Scratch receives the same deployment budget without meta-training. Both phases use Adam
at $2.5\times10^{-4}$ and 1,024 fresh interior points per step; evaluation occurs every
100 deployment steps. The weighted residual loss uses $(w_N,w_\Phi)=(0.003,0.997)$.
The orthogonality penalty is added once after combining residual losses, with weight
$10^{-4}$ during meta-training and zero during deployment.
The coefficient sampler uses $V_0\in[130,270]$ and logarithmic
$R_0\in[10^{20},5\times10^{20}]$, holding its centre-out radius at zero through step
750 and reaching the full box at step 2,250. The operator graph applies two non-residual update rounds.

\FloatBarrier
\section{Supplementary comparisons}
\label{app:comparisons}
\subsection{CDR: term composition and seed variation}
\label{app:cdr-evidence}
\label{app:cdr-seeds}
\begin{figure}[htbp]
\centering
\includegraphics[width=0.93\linewidth]{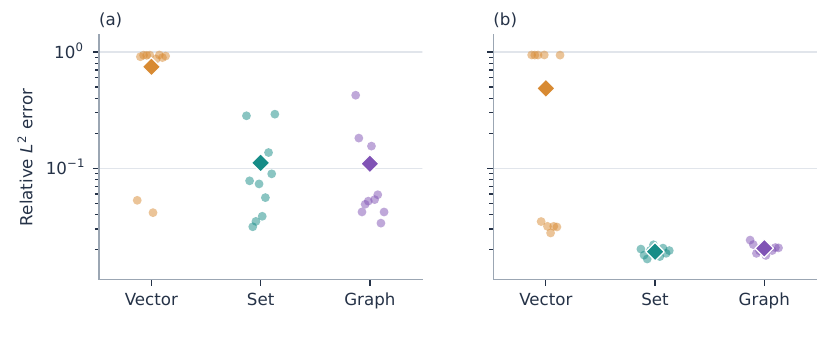}
\caption{CDR oracle-best error: five-to-two (left) and all-seven training (right).
Circles average eight AR/ADR cases within each seed; diamonds average ten seeds.}
\label{fig:cdr-high-seeds}
\end{figure}

Figure~\ref{fig:cdr-high-seeds} complements Table~\ref{tab:cdr-ten-seed} with the
ten seed means. Set and Graph perform similarly in both high-reaction protocols;
including AR/ADR during meta-training lowers their errors and seed variation.

\paragraph{Fixed-budget endpoints.}
Table~\ref{tab:cdr-final} reports the step-3000 errors for the same ten seeds and
eight AR/ADR cases as Table~\ref{tab:cdr-ten-seed}, without checkpoint selection.
Both term-based descriptors retain substantially lower mean errors than Vector.
All-seven training again lowers their errors and seed variation; Graph does not
consistently outperform Set across the two protocols.

\begin{table}[htbp]
\centering
\caption{High-reaction CDR final $\rell$ at step 3000: mean $\pm$ sample SD
over ten seed means, each averaging the same eight cases as Table~\ref{tab:cdr-ten-seed}.}
\label{tab:cdr-final}
\small
\begin{tabular}{@{}lccc@{}}
\toprule
Meta-training & Coefficient vector & Term set & Operator graph \\
\midrule
Five-to-two & $0.7682\pm0.3800$ & $0.1625\pm0.1775$ & $0.1382\pm0.1590$ \\
All seven   & $0.4909\pm0.4842$ & $0.0201\pm0.0015$ & $0.0210\pm0.0017$ \\
\bottomrule
\end{tabular}
\end{table}

In the lower-reaction five-to-two protocol of App.~\ref{app:cdr-ten-seed}, the
oracle-best means are $0.0210\pm0.0075$ (Vector), $0.0119\pm0.0031$ (Set), and
$0.0312\pm0.0233$ (Graph), each averaging six cases within a seed and then ten seeds.
Together, these comparisons support term-based conditioning in high-reaction CDR,
with no consistent Graph advantage over Set across the evaluated regimes.

\FloatBarrier
\Needspace{10\baselineskip}
\subsection{Fisher--KPP: accuracy and coefficient-presence control}
\label{app:fisher-evidence}
Table~\ref{tab:fisher-aggregate} reports final and oracle-best errors for the
five-seed coupling-composition experiment. The extra vector control adds presence
indicators to distinguish an absent coupling from a mean-valued active coefficient.
It improves the mean error relative to the unmasked vector, while the operator graph
retains the lowest mean error.

\begin{table}[htbp]
\centering
\caption{Fisher--KPP relative errors in units of $10^{-3}$: mean $\pm$ sample SD
over five seed means. Final uses step 3000; oracle-best uses each case's lowest scheduled error.}
\label{tab:fisher-aggregate}
\small
\begin{tabular}{@{}lcc@{}}
\toprule
Method & Final & Oracle-best \\
\midrule
From scratch & $9.660\pm0.626$ & $9.660\pm0.626$ \\
Coefficient vector & $5.531\pm1.302$ & $5.348\pm1.458$ \\
Term set & $2.779\pm0.712$ & $2.754\pm0.698$ \\
Operator graph & $1.786\pm0.318$ & $1.658\pm0.410$ \\
\addlinespace
Vector with presence indicators & $4.191\pm2.174$ & $3.884\pm2.267$ \\
\bottomrule
\end{tabular}
\end{table}

\paragraph{Consistency across held cases and fields.}
Table~\ref{tab:fisher-per-case} separates the same five-seed cohort by held instance.
Graph has lower mean final error than Set and Vector on each of H1--H4.
Across matched case--seed pairs, Graph is lower than Set in 18/20 and lower than
Vector in 20/20; these pairs comprise five independent training seeds, not 20.
The Graph--Set ordering also holds separately for each field after averaging the
four cases within each seed: the five-seed means are $1.306$ versus $2.233$ for $u$
and $2.266$ versus $3.326$ for $v$, all in units of $10^{-3}$.
Thus the aggregate improvement is not confined to one held case or one field.

\begin{table}[htbp]
\centering
\small
\begin{tabular}{@{}lccc@{}}
\toprule
Case & Coefficient vector & Term set & Operator graph \\
\midrule
H1 & $3.142\pm0.599$ & $1.640\pm0.445$ & $1.044\pm0.218$ \\
H2 & $3.298\pm0.491$ & $1.785\pm0.408$ & $1.122\pm0.308$ \\
H3 & $7.478\pm2.653$ & $3.735\pm0.458$ & $1.936\pm0.280$ \\
H4 & $8.204\pm2.019$ & $3.957\pm1.815$ & $3.043\pm0.887$ \\
\bottomrule
\end{tabular}
\normalsize
\caption{Fisher--KPP final error by held case, in units of $10^{-3}$.
Mean $\pm$ sample SD over five seeds, averaging the two fieldwise errors.}
\label{tab:fisher-per-case}
\end{table}

\FloatBarrier
\subsection{Fisher--KPP: ownership-aware term conditioning}
\label{app:fisher-ownership}
To test whether explicit field/equation assignments suffice for the term set,
we add two target-equation indicators and eight source-factor multiplicities to
its six-attribute token. This 16-attribute token uses the same DeepSets pooling
and field-conditioned readout, without graph adjacency.
The control uses seeds 101, 202, and 303 with the backbone, budgets, and
within-seed coefficient/collocation streams of App.~\ref{app:fisher}.
These seed IDs overlap the five-seed comparison.

\begin{table}[htbp]
\centering
\caption{Ownership control: final relative error at step 3000, in units of $10^{-3}$.
Mean $\pm$ sample SD over three seed means, each averaging H1--H4 and both fields.}
\label{tab:fisher-ownership}
\small
\begin{tabular}{@{}lc@{}}
\toprule
Method & Final \\
\midrule
Term set & $3.065\pm0.453$ \\
Ownership-aware term set & $4.829\pm1.951$ \\
Operator graph & $1.687\pm0.136$ \\
\bottomrule
\end{tabular}
\end{table}

Adding ownership attributes does not close the gap to the graph under this budget.
This comparison concerns complete encoders, including their processing and readouts.

\clearpage
\subsection{Fisher--KPP: message-passing depth}
\label{app:fisher-depth}
We test whether the graph-conditioned initialization and its subsequent adaptation
benefit from additional message-passing rounds. This ablation varies the recurrence
count $K\in\{1,2,3,4,5\}$ in the encoder of \S\ref{sec:method}, retaining the
operator graph, node features, readout, and PINN backbone. The same $K$ is used in
meta-training and when encoding a held equation. Because the update weights are
shared across rounds, all depths have 132,866 parameters.

\paragraph{Matched protocol.}
We use the Fisher--KPP split, coefficient distribution, hard initial/periodic
constraints, and optimization settings of App.~\ref{app:fisher}. Seeds
$101,202,303,404,505$ each define five depth models with identical initial
parameter tensors and coefficient/collocation streams within the seed. Every
model receives 3,000 meta-training updates, followed by 3,000 full-adaptation
updates on each of H1--H4, starting from a fresh model copy for every case.
The $K=2$ setting matches the main graph comparison. We evaluate initialization
and steps 500 and 3000, first averaging relative errors over both fields and four
held cases within a seed, then reporting the mean and sample SD across seeds.

\begin{table}[htbp]
\centering
\caption{Matched-depth Fisher--KPP ablation. Relative $L_2$ error in units of
$10^{-3}$: mean $\pm$ sample SD over five seed means.}
\label{tab:fisher-depth}
\small
\begin{tabular}{@{}lccc@{}}
\toprule
Rounds $K$ & Initialization & Step 500 & Step 3000 \\
\midrule
1 & $143.86\pm7.80$ & $11.583\pm2.063$ & $2.691\pm0.971$ \\
2 & $138.40\pm6.76$ & $10.107\pm1.572$ & $1.786\pm0.318$ \\
3 & $131.95\pm5.79$ & $8.822\pm1.025$ & $1.704\pm0.226$ \\
4 & $136.70\pm12.28$ & $8.130\pm1.384$ & $1.572\pm0.240$ \\
5 & $109.39\pm23.32$ & $6.118\pm1.136$ & $1.311\pm0.352$ \\
\bottomrule
\end{tabular}
\end{table}

\begin{figure}[htbp]
\centering
\includegraphics[width=\linewidth]{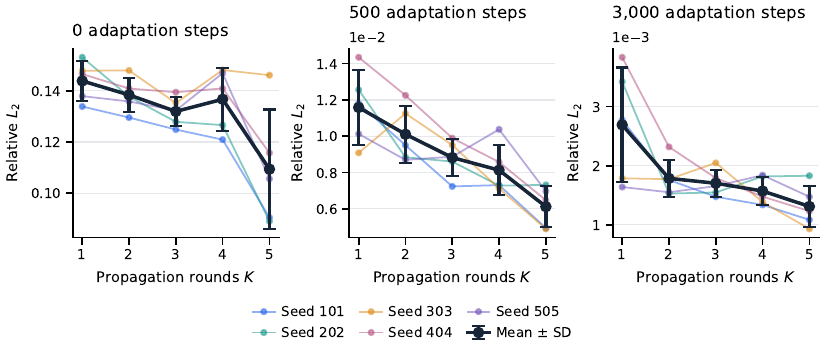}
\caption{Depth sensitivity at initialization and two adaptation checkpoints.
Thin lines connect the five depth models within each seed; the bold line and
whiskers show the mean and sample SD. Training and test depths are matched.
Vertical scales differ across checkpoints.}
\label{fig:fisher-depth}
\end{figure}

\paragraph{Depth sensitivity.}
Table~\ref{tab:fisher-depth} and Fig.~\ref{fig:fisher-depth} show lower mean error
at five rounds than at two: reductions of $21.0\%$ at initialization, $39.5\%$
after 500 updates, and $26.6\%$ after 3000. The paired five-versus-two comparison
improves all five seeds at the first two checkpoints and four of five at the last;
seed 202 has lower final error at two rounds. Intermediate depths do not improve
monotonically in every seed. Thus additional graph processing can improve
fixed-budget adaptation at unchanged parameter count, while the benefit varies
across training runs. This comparison holds optimizer updates fixed, not compute
time: deeper encoding requires more computation.

\Needspace{5\baselineskip}
As a separate equation-consistency diagnostic, we measure mean squared residual
over both equations on a fixed $32\times32$ midpoint grid, then average H1--H4
within each seed. At step 3000, residual MSE decreases from
$(6.907\pm2.565)\times10^{-5}$ at $K=2$ to
$(3.623\pm0.994)\times10^{-5}$ at $K=5$. This diagnostic uses no reference
solution values and is distinct from the relative solution error above.

\begin{figure}[!htbp]
\centering
\includegraphics[width=\linewidth]{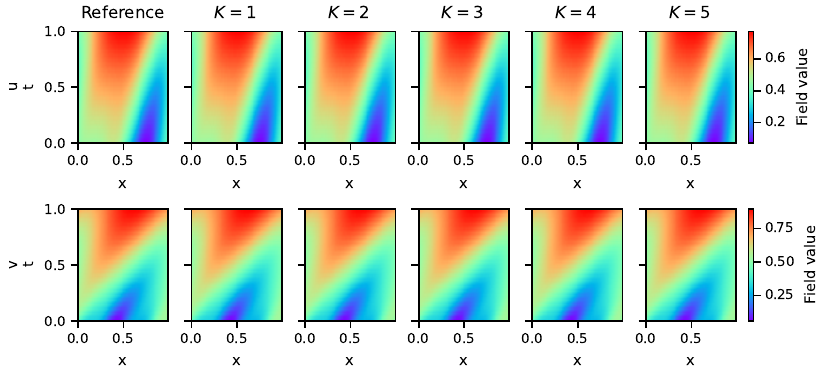}
\par\medskip
\includegraphics[width=\linewidth]{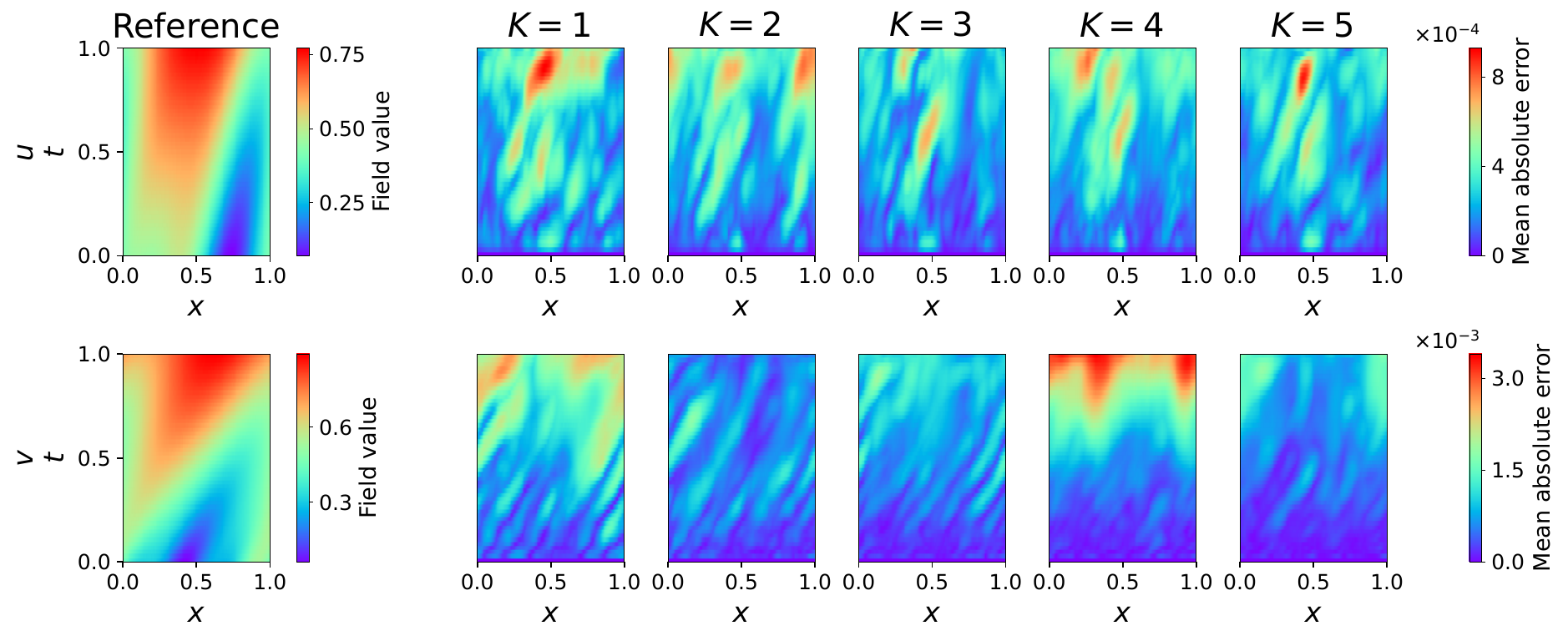}
\caption{H1 after 3000 adaptation steps. Top: reference and predicted fields for
the predeclared seed 101, with one shared field-value scale per row. Bottom:
reference fields and the five-seed mean of pointwise absolute errors, computed
before averaging seeds. Each pair of rows shows $u$ and $v$. In the bottom panel,
the reference colorbar follows the left column; the error colorbar is at the far right.
All maps use linear rainbow scales, shared across depths within each field and
quantity. Display grids contain 128 spatial and 51 temporal points.}
\label{fig:fisher-depth-fields}
\end{figure}

Figure~\ref{fig:fisher-depth-fields} complements the aggregate comparison with
the H1 fields. The predictions recover the main spatiotemporal structure at every
depth, while the error maps reveal remaining local differences. Scalar results
in Table~\ref{tab:fisher-depth} use the full reference grid and all four held cases.

\clearpage
\section{Coefficient interpolation under different training samplers}
\label{app:coefficient-sampling}
The all-seven CDR comparison in \S\ref{sec:exp-cdr} shows that term-based descriptors
can configure accurate solvers at new coefficients of familiar structures. Its
meta-training sampler gradually expands around coefficient-grid nodes. This raises
a complementary question: does the observed accuracy require that particular
sampling schedule, or does it persist when training sees either the full continuous
range from the start or only the discrete grid? We address this question by fixing
the training structures and varying the coefficient sampler. The comparison concerns
coefficient interpolation after physics-based adaptation, rather than new term
combinations.

\subsection{Controlled coefficient-interpolation setting}
All three sampling conditions train on the seven CDR structures in
Table~\ref{tab:structures}. They share the initial and boundary conditions, reference
solver, descriptor definitions, and coefficient grid of
Apps.~\ref{app:family} and~\ref{app:cdr-descriptors}. For each active axis, a grid
node is sampled and perturbed within a fraction $\alpha(m)$ of its half-spacing
at meta-step $m$, with clipping at the axis limits. Diffusion is sampled in
$\log_{10}D$. The three conditions are:
\begin{itemize}\itemsep2pt
\item \emph{Scheduled jitter}: $\alpha(m)=0$ through step 8,000, then increases
linearly to one by step 24,000 and remains one through step 30,000.
\item \emph{Full jitter}: $\alpha(m)=1$ throughout training, so the continuous
coefficient box is accessible from the first step.
\item \emph{Grid-only}: $\alpha(m)=0$ throughout training, so only the discrete
nodes are accessible.
\end{itemize}
At full radius, neighboring perturbation intervals meet; the clipped sampling
distribution need not be uniform. Scheduled and full jitter have the same eventual
support but different exposure to intermediate values during training. Grid-only
retains the range endpoints and all grid nodes while removing the intervening
continuous support.

\paragraph{Shared optimization protocol.}
This supplementary experiment uses the width-256 backbone with three coded layers,
two graph-update rounds, 30,000 meta-training steps, and 3,000 full-adaptation steps.
Adam uses learning rate $2.5\times10^{-4}$ in both phases. Each loss evaluation
resamples 20,000 interior points, 1,000 initial points, and 1,000 periodic-boundary
pairs. Residual, initial, and periodic-value losses have unit weights; the
basis-orthogonality weight is $10^{-4}$ during meta-training and zero during
adaptation. Meta-training averages the losses of all
seven structures and adds the basis penalty once. Descriptor preprocessing and
architecture are held fixed across samplers. These larger collocation batches
distinguish this experiment from the ten-seed main comparison; its results are
reported separately.

\paragraph{Evaluation cases and aggregation.}
The 12 high-reaction cases comprise reaction and diffusion--reaction at
$r\in\{8.5,9.5\}$, and AR/ADR at
$a\in\{4.5,9.5\}\times r\in\{8.5,9.5\}$. Diffusion-bearing cases use $D=0.01$.
Both test reaction rates lie strictly between adjacent training nodes $7.75$ and
$10$, while the test advection and diffusion values are grid nodes. Thus grid-only
deployment specifically tests interpolation in reaction strength, without changing
the equation structures or extrapolating beyond the coefficient ranges.

We evaluate every 200 adaptation steps and use each case's oracle-best
$\rell$ within the budget, including initialization. Table~\ref{tab:sampler-errors}
averages the 12 case errors within each seed and then across seeds. Its hit fraction
is the proportion of case--seed deployments with oracle-best $\rell<0.3$; reference
fields enter only this evaluation, not training or online stopping. Scheduled and
full jitter use three seeds per descriptor. Grid-only uses six seeds for Set and
Graph and three for Vector. Each seed contributes the same 12 cases within a
condition, and unsuccessful deployments remain in both the error and hit summaries.

\Needspace{10\baselineskip}
\subsection{Gradual widening is not required for low error}
Table~\ref{tab:sampler-errors} compares accuracy under the three samplers.
With scheduled jitter, Set and Graph both have mean error $0.010$ and every
evaluated deployment reaches the stated threshold. Full jitter gives closely
similar errors, $0.013$ for Set and $0.012$ for Graph, with the same hit fractions.
Their low errors therefore persist when the continuous coefficient range is
available from the beginning. The gradual-widening schedule is not required for
this outcome under the tested budget.

\begin{table}[htbp]
\centering
\caption{CDR coefficient-sampling comparison: mean oracle-best $\rell$ and hit
fraction ($\rell<0.3$). Scheduled/full jitter use three seeds per method;
grid-only uses three Vector and six Set/Graph seeds, with 12 cases per seed.}
\label{tab:sampler-errors}
\small
\begin{tabular}{@{}l cc cc cc@{}}
\toprule
& \multicolumn{2}{c}{Scheduled jitter}
& \multicolumn{2}{c}{Full jitter}
& \multicolumn{2}{c}{Grid-only} \\
\cmidrule(lr){2-3}\cmidrule(lr){4-5}\cmidrule(lr){6-7}
Descriptor & Error & Hit & Error & Hit & Error & Hit \\
\midrule
Coefficient vector & $0.958$ & $0.00$ & $0.330$ & $0.67$ & $0.958$ & $0.00$ \\
Term set           & $0.010$ & $1.00$ & $0.013$ & $1.00$ & $0.200$ & $0.79$ \\
Operator graph     & $0.010$ & $1.00$ & $0.012$ & $1.00$ & $0.010$ & $1.00$ \\
\bottomrule
\end{tabular}
\end{table}

The coefficient vector also responds to the sampler: its mean error falls from
$0.958$ with scheduled jitter to $0.330$ with full jitter, and its hit fraction
increases from zero to $0.67$. It still has higher error than the term-based
descriptors. This comparison shows that sampling choices can change the size of a
descriptor gap; the term-based advantage in this experiment is nevertheless
present under both continuous-support conditions.

\subsection{Grid-only training exposes seed variation}
Removing continuous coefficient support separates Set and Graph in this cohort.
The graph's mean error remains $0.010$, while Set's mean increases to $0.200$.
The seed-level counts in Table~\ref{tab:sampler-seeds} show why the aggregate
gap should not be read as uniformly worse Set performance: four of its six
training seeds reach the threshold on all 12 cases, one reaches it on nine,
and one on none. Graph reaches it on all 12 cases for each of six seeds.

\begin{table}[htbp]
\centering
\caption{Grid-only deployments with oracle-best $\rell<0.3$, out of 12 per seed.
The total combines deployments, not independent training runs.}
\label{tab:sampler-seeds}
\small
\begin{tabular}{@{}lccccccc@{}}
\toprule
Descriptor & Seed 0 & Seed 1 & Seed 2 & Seed 3 & Seed 4 & Seed 5 & Total \\
\midrule
Coefficient vector & $0/12$ & $0/12$ & $0/12$ & --- & --- & --- & $0/36$ \\
Term set           & $12/12$ & $0/12$ & $9/12$ & $12/12$ & $12/12$ & $12/12$ & $57/72$ \\
Operator graph     & $12/12$ & $12/12$ & $12/12$ & $12/12$ & $12/12$ & $12/12$ & $72/72$ \\
\bottomrule
\end{tabular}
\end{table}

Within the shared seed IDs 0--2, the counts are $36/36$ for Graph, $21/36$
for Set, and $0/36$ for Vector. In the additional seeds 3--5, both Set and Graph
reach $36/36$. The pooled hit-fraction gap therefore comes from two of the six
Set training seeds, rather than a loss of accuracy in every seed.
Seed IDs are shared across descriptors, but
different architectures do not produce pointwise-matched collocation streams.

\paragraph{Implication and scope.}
Together, these controls support two specific observations: low Set/Graph error
persists without gradual widening, and Graph is more consistent across the sampled
seeds when meta-training sees only grid coefficients. Both observations concern
solver adaptation to intermediate reaction rates within familiar equation structures
and a bounded coefficient range. Each target solver receives 3,000 physics-based
updates. These results complement the coupling-composition experiment by showing
how numerical coefficient exposure affects adaptation under the three conditioning
schemes.

\clearpage
\section{Optimization steps and amortization}
\label{app:amortization}

\paragraph{Reaching a common solution accuracy.}
A separate lower-reaction \CDR{} experiment measures the optimization head-start
provided by equation-conditioned initialization. Five structures are used for
meta-training, with advection--reaction and advection--diffusion--reaction held
out. Training samples $a\in[2,4]$, $r\in[1,4]$, and $\log_{10}D\in[-3,-1]$
for active terms. Each method uses three independent seeds, a width-256 backbone
with three coded layers, and full target-specific adaptation. The amortized
methods receive 20,000 meta-training updates, each averaging one sampled problem
per training structure. Physics losses use 25,000 interior, 1,000 initial, and
1,000 periodic samples, with Adam at $2.5\times10^{-4}$. Deployment runs for
5,000 updates, evaluating the reference-grid error every 200 updates, including
initialization. This cohort is separate from the ten-seed comparisons in
App.~\ref{app:cdr-ten-seed}.

For each deployment, $n_{0.1}$ is the first scheduled update with $\rell\leq0.1$.
Table~\ref{tab:cdr-amortization} reports its median among successful deployments,
together with the fraction reaching the threshold. The control contains ten
held-coefficient cases per seed on seen structures; the structural test contains
six cases per seed on the two held structures. Reference errors determine these
evaluation statistics only; they are not used for training or online stopping.

\begin{table}[htbp]
\centering
\small
\begin{tabular}{@{}lrrrr@{}}
\toprule
& \multicolumn{2}{c}{Seen structure, held coefficients}
& \multicolumn{2}{c}{Held structure}\\
\cmidrule(lr){2-3}\cmidrule(lr){4-5}
Method & Median $n_{0.1}$ & Reach fraction & Median $n_{0.1}$ & Reach fraction\\
\midrule
From scratch       & 600 & $90\%$ & 2200 & $94\%$\\
Coefficient vector & 200 & $100\%$ & 800 & $100\%$\\
Term set           & 200 & $100\%$ & 400 & $100\%$\\
Operator graph     & 200 & $100\%$ & 600 & $100\%$\\
\bottomrule
\end{tabular}
\normalsize
\caption{\CDR{} updates to $\rell\leq0.1$: three seeds, with 30 control and
18 held-structure deployments per method. Medians condition on threshold
attainment; reach fractions are rounded.}
\label{tab:cdr-amortization}
\end{table}

On held structures, the graph requires 600 rather than 2200 median updates,
giving a ratio of $3.67$. The vector and set also reduce the required updates.
Across seeds, the mean and population standard deviation of the within-seed
medians are $2267\pm170$ for scratch, $833\pm125$ for Vector, $467\pm125$ for
Set, and $633\pm340$ for Graph. These results support an amortized optimization
head-start, while the Fisher--KPP comparison in \S\ref{sec:exp-fisher} addresses
the additional value of relational conditioning.

These are update-count comparisons: a grouped meta-training update processes
five PDE instances, whereas a deployment update processes one. Matched-device
Fisher--KPP timings are reported separately in App.~\ref{app:fisher-iwata}.

\clearpage
\section{Conditional-decoder comparison}
\label{app:fisher-iwata}

\subsection{A matched-budget external method}
\label{app:fisher-iwata-protocol}
The descriptor comparison in \S\ref{sec:exp-fisher} fixes the solver backbone.
Here we compare the complete graph-conditioned solver with a conditional
decoder following \citet{iwata2023meta}, whose equation and condition encoders
produce a latent input to a sinusoidal solution network. Its deployment phase
refines a copied decoder and this latent while keeping the encoder fixed.
Both approaches therefore share physics information across equations and
retain target-specific refinement, but use different solver architectures.

We extend the polynomial coefficient dictionary to the two Fisher fields.
For each residual, the variables
$(u,u_t,u_x,u_{tt},u_{tx},u_{xx},v,v_t,v_x,v_{tt},v_{tx},v_{xx})$
give one constant, 12 linear and 78 quadratic monomials. Concatenating the
two residual rows gives 182 signed coefficients, with absent monomials
represented by zero. The condition encoder pools 64 prescribed initial
point--value pairs; the initial profiles are fixed across this benchmark.
Four-layer ReLU networks encode coefficients and condition points, followed
by pooling and a linear combination layer. A coordinate projection and the
latent feed a five-layer sine decoder with residual hidden connections.

All methods use the Fisher task distribution, periodic coordinate features,
hard initial-condition construction, physics objective and optimizer budgets
of App.~\ref{app:fisher}. The conditional decoder uses hidden and latent width
82, chosen by parameter count: 132,514 parameters versus 132,866 for Graph.
Its decoder has 34,934 parameters; the remainder belongs to the encoder.
Linear layers use the PyTorch default initialization. We refer to this
two-field, parameter-matched implementation as \emph{Iwata-style}: the source
paper uses scalar equations, Dirichlet conditions, width 256 and a different
training distribution. This comparison evaluates complete methods at a
matched total parameter budget, alongside the fixed-backbone descriptor
controls in the main text.

The external-method cohort runs Iwata-style, two-round Graph and scratch
serially on one NVIDIA GB10 with PyTorch 2.10.0a0 and CUDA 13, float32,
without AMP or TF32. Each method uses seeds 101, 202, 303, 404 and 505,
the same per-seed coefficient and collocation streams, 3,000 grouped
meta-updates for the trained methods, and 3,000 adaptation updates on
each of H1--H4. All comparisons in this section use this cohort; the CPU
descriptor results in the main text remain a separate experiment.

\begin{table}[htbp]
\centering
\small
\begin{tabular}{@{}lccc@{}}
\toprule
Method & Initialization & Step 500 & Step 3000 \\
\midrule
From scratch & $544.18\pm63.46$ & $84.86\pm3.81$ & $9.66\pm0.63$ \\
Iwata-style & $88.10\pm14.43$ & $12.84\pm0.76$ & $3.27\pm0.31$ \\
Operator graph & $138.04\pm6.29$ & $9.80\pm1.38$ & $1.81\pm0.30$ \\
\bottomrule
\end{tabular}
\normalsize
\caption{Matched-device Fisher--KPP external-method comparison. Relative
$L_2$ errors in units of $10^{-3}$: mean $\pm$ sample SD over five seed means.}
\label{tab:fisher-iwata-accuracy}
\end{table}

Table~\ref{tab:fisher-iwata-accuracy} shows lower initialization error for
Iwata-style, but lower Graph error after 500 and 3000 updates. At the final
checkpoint, Graph reduces the mean by $44.6\%$ relative to Iwata-style,
with lower case-averaged error in all five seeds. Thus the ordering of
initial predictions differs from that after physics refinement.

\paragraph{Measured solve cost.}
Meta-training takes $356.1\pm16.5$ seconds for Graph and $285.9\pm17.0$
seconds for Iwata-style (mean $\pm$ sample SD across five seeds).
All methods reach $\rell\leq0.1$ on all 20 deployments; Graph's mean online
time to the first scheduled hit is $4.12\pm0.24$ seconds per problem,
versus $16.54\pm1.84$ from scratch, while Iwata-style has lower online cost
than Graph at this loose target.
CUDA-synchronized online time includes encoding, target setup and optimization,
but excludes reference generation and diagnostic evaluation.
Threshold hits are evaluated every 100 updates, including initialization,
retrospectively over the full 3,000-update trajectories rather than used
as an online stopping rule.

\end{document}